\documentclass{article}

\usepackage[main, final]{neurips_2026}

\usepackage[utf8]{inputenc} % allow utf-8 input
\usepackage[T1]{fontenc}    % use 8-bit T1 fonts
\usepackage{hyperref}       % hyperlinks
\usepackage{url}            % simple URL typesetting
\usepackage{booktabs}       % professional-quality tables
\usepackage{amsfonts}       % blackboard math symbols
\usepackage{nicefrac}       % compact symbols for 1/2, etc.
\usepackage{microtype}      % microtypography
\usepackage{xcolor}         % colors

\usepackage{graphicx}
\usepackage{subcaption}

\usepackage{amsmath}
\usepackage{amssymb}
\usepackage{mathtools}
\usepackage{amsthm}
\usepackage{algorithm}
\usepackage{algpseudocode}
\usepackage{multirow} 
\usepackage{wrapfig}
\usepackage{bbm}

\theoremstyle{plain}
\newtheorem{theorem}{Theorem}[section]

\newtheorem{lemma}[theorem]{Lemma}

\theoremstyle{definition}

\theoremstyle{remark}
\newtheorem{remark}[theorem]{Remark}

\newcommand{\subfigH}{4.9cm}

\title{BReD: Block Replay Dithering for Stable Low-Bit EMA Optimizer States}

\author{%
  \textbf{Heshen Zhan}$^{1,4}$,
  \textbf{Youhan Huang}$^{1}$,
  \textbf{Yunke Peng}$^{2}$,
  \textbf{Yao Wang}$^{2}$,
  \textbf{Linghui Kong}$^{2}$\\[0.5em]
  \textbf{Ziwei Zhu}$^{1}$,
  \textbf{Qingyu Han}$^{1,4}$, 
  \textbf{Yaoyuan Wang}$^{2}$,
  \textbf{Congliang Chen}$^{3,4,\dagger}$,
  \textbf{Ruoyu Sun}$^{1,3,4,\dagger}$ \\
  $^{1}$The Chinese University of Hong Kong, Shenzhen \\
  $^{2}$Huawei \\
  $^{3}$Shenzhen Loop Area Institute \\
  $^{4}$SICIAM, Shenzhen Research Institute of Big Data
}

\begin{document}
\maketitle

\begingroup
\renewcommand{\thefootnote}{\ensuremath{\dagger}}
\footnotetext{Corresponding authors:
  Ruoyu Sun (\texttt{sunruoyu@cuhk.edu.cn}) ;
  Congliang Chen (\texttt{chencongliang@slai.edu.cn}).}
\endgroup

\begin{abstract}

Low-bit storage of optimizer states can substantially reduce the memory footprint of large-scale training. In practice, however, these states are not quantized once: at every training step, they are dequantized, updated, requantized, and written back. This iterative quantization pipeline can destabilize training. We identify a failure mode behind this instability: for states updated via exponential moving averages (EMAs), quantization errors are recursively fed into subsequent updates. As a result, biased errors accumulate over time while high-variance errors are amplified, particularly when the EMA decay factor is close to 1. To address this error accumulation, we propose BReD (\textbf{B}lock \textbf{Re}play \textbf{D}ithering), a method that reduces rounding bias and controls rounding variance during quantization. BReD adapts classical subtractive dithering to optimizer-state quantization by combining deterministic seed replay with block-wise shared dither values, avoiding auxiliary random-tensor storage and per-element dither values generation. Across the default pretraining settings (five optimizers at 120M; AdamW and Muon at 1.1B and 3.4B model) and supervised fine-tuning (AdamW and Muon at 7B model), 4-bit BReD closely matches training with full-precision optimizer states, with PPL shifts $\leq0.5$ in pretraining and $\leq0.2$ in fine-tuning and average downstream performance changes within 0.7 points. Moreover, 3-bit BReD preserves stable convergence in the evaluated settings, offering a more aggressive alternative.
\footnote{Our implementation of BReD is available at: \url{https://github.com/ZhanHeshen/BReD}.}

\end{abstract}
  
\section{Introduction}
In large-scale model training, many optimizers maintain auxiliary states, such as moment estimates or preconditioner states.
For example, AdamW \citep{kingma2014adam,loshchilov2017decoupled} keeps first- and second-moment estimates, Muon \citep{jordan6muon} uses a momentum buffer, and SOAP \citep{vyas2024soap} maintains matrix preconditioner states.
These auxiliary states are used to compute the optimizer update, but they also introduce additional memory cost because they must be kept throughout training. 
Since this cost increases with model size, storing optimizer states in low-bit formats is a direct way to lower the memory footprint of training. 

A growing line of work has studied low-bit optimizer-state storage as a way to reduce this cost, including low-bit Adam variants \citep{li2023memory,xu2025pushing}, low-bit Muon methods~\citep{wuachieving,gupta2025effective}, and low-bit Shampoo methods~\citep{wang20244,li2025memory}. These methods show that optimizer-state memory can be substantially reduced. However, at aggressive low bitwidths, stability becomes harder to maintain, and successful recipes are often tied to particular optimizer families, update modifications, or retuned hyperparameters. 
This raises a complementary question: can low-bit optimizer-state storage be stabilized across different optimizers while preserving the original optimizer update and hyperparameters?

We address this question by shifting focus from optimizer update rules to a shared mechanism in many optimizers: exponential moving average (EMA) recursions. 
For EMA optimizer states, low-bit storage is not a one-time quantization problem.
At each training step, the low-bit state is dequantized for the optimizer update, then requantized and written back for future use. 
% This repeated writeback is especially consequential for states updated by exponential moving average (EMA) recursions, \(s_{t+1}=\beta s_t+(1-\beta)u_{t+1}\), where \(\beta\) is the EMA decay factor. 
Let \(e_t\) denote the difference between the dequantized stored state and the corresponding full-precision state, 
and let \(\epsilon_{t+1}\) be the error injected when the updated state is quantized. 
Then \[e_{t+1}=\beta e_t+\epsilon_{t+1},\] 
where \(\beta\) is the EMA decay factor (Sec.~\ref{sec:ema_feedback}).
Thus, the effect of quantization is governed not only by the error injected at each step, 
but also by how this error accumulates through the EMA recursion. 

% This recursion points to 
The rounding rule is a controllable component of quantization. 
Ignoring clipping or saturation, quantization error is determined by the rounding rule for a given scale and quantization grid.
The error accumulation points to three design goals for the rule.
First, rounding error should be unbiased, as even small systematic biases can accumulate into persistent drift. 
Second, unbiasedness alone is not enough: the error variance should also be controlled, because the EMA recursion amplifies the variance term as well. 
Third, the rule must be practical at training scale: optimizer states may contain billions of entries, so it should not require storing large random tensors or generating independent random numbers for every element.

Guided by these design goals, we propose \textbf{BReD} (\textbf{B}lock \textbf{Re}play \textbf{D}ithering), a rounding-based scheme for low-bit storage of EMA optimizer states. 
BReD applies subtractive dithering ~\citep{roberts1962picture} in quantization and dequantization, producing unbiased reconstruction error with controlled variance. 
While subtractive dithering is classical, applying it to optimizer-state storage is nontrivial: identical dither values must be available at the dequantization time, which typically requires storing large random tensors or regenerating dither values for every element.
BReD addresses this challenge using deterministic seed replay and block-wise sharing of dither values.
It enables practical subtractive dithering in large-scale training without modifying optimizer updates or hyperparameters.

We evaluate BReD on pretraining and supervised fine-tuning. The pretraining experiments cover five optimizers at 120M model, AdamW/Muon at 1.1B and 3.4B model and a long-horizon AdamW run, while the fine-tuning experiments evaluate AdamW and Muon at 7B model. Across these settings, 4-bit BReD 
incurs small changes relative to the corresponding full-precision baselines,
with perplexity shifts within \(0.5\) in pretraining and \(0.2\) in fine-tuning, and average downstream performance changes within \(0.7\) points. This stability is not simply a consequence of the low-bit format: under the same scaling rule and quantization grid, replacing BReD with nearest or stochastic rounding can diverge or downgrade during pretraining. Moreover, 3-bit BReD remains stable in the evaluated settings, offering an aggressive alternative.

Our contributions are:
\begin{itemize}
    \item \textbf{Mechanism.} We identify error accumulation as a critical failure mode behind low-bit storage of EMA optimizer states, and we characterize its underlying mechanism: the bias and variance of quantization errors continuously propagate through EMA recursive updates.

    \item \textbf{Method.} We propose BReD, a quantization scheme for low-bit storage of EMA optimizer states. It adapts classical subtractive dithering with two designs: deterministic seed replay and block-wise shared dither values. Importantly, BReD leaves the original optimizer update rules and hyperparameter configurations unchanged.

    \item \textbf{Evaluation.} We evaluate BReD across various optimizers, model scales, and training setups. Our experiments include rounding baselines, low-bit optimizer methods, a longer-horizon test, and memory-perplexity Pareto analysis. BReD enables robust 4-bit optimizer state compression in all evaluated settings, and remains convergent at 3-bit precision in the evaluated settings.
\end{itemize}

\section{Related Work}
\paragraph{Low-bit optimizer states.}
Optimizer states are a major memory bottleneck in large-scale training. Existing approaches reduce this cost structurally, as in Adafactor~\citep{shazeer2018adafactor} and Adam-mini ~\citep{zhang2024adam} or at the system level, as in ZeRO-style sharding/offloading ~\citep{rajbhandari2020zero,ren2021zero,rajbhandari2021zero}. We study a complementary axis: low-bit storage of EMA optimizer states. Prior work has shown strong results for block-wise 8-bit optimizers ~\citep{dettmers20218} and 4-bit AdamW ~\citep{li2023memory}; we refer to the latter as Adam-LPMM. 
Recent methods further push low-bit AdamW states by focusing
on EMA dynamics, as in SOLO~\citep{xu2025pushing}, or target
Muon states~\citep{wuachieving,gupta2025effective} and
preconditioner states, as in 4-bit Shampoo~\citep{wang20244}
and 4-bit Preconditioned Optimization~\citep{li2025memory}.
% Recent methods further push low-bit AdamW states by focusing on EMA dynamics \citep{xu2025pushing}, or use optimizer-specific structure for Muon- and Shampoo-style states ~\citep{wuachieving,gupta2025effective}. 
In contrast, BReD targets the shared quantize--write-back path of EMA states. It changes the rounding and reconstruction rule used for state storage, while keeping the optimizer update and hyperparameters unchanged.

\paragraph{Unbiased rounding and dithering.}
Unbiased rounding reduces systematic rounding bias. Stochastic rounding is unbiased in expectation ~\citep{von1947numerical,alistarh2017qsgd}. Subtractive dithering is a classical technique that gives unbiased reconstruction ~\citep{roberts1962picture,jayant1972application,1088973,sherwood1985some}; under the local non-clipped interval model used in \ref{sec:round}, it has smaller worst-case conditional variance than stochastic rounding. However, subtractive dithering requires the same dither value at quantization and dequantization time, which is impractical at LLM scale if per-element dither values must be stored or generated. BReD makes subtractive dithering practical for persistent EMA optimizer states through deterministic seed replay and block-wise shared dither values.

\section{Understanding Instability: Quantized EMA Error Accumulation}
\label{sec:ema_error}
Many optimizer states follow EMA recursions, such as the first and second moments in AdamW and the momentum buffer in Muon.
When these states are quantized and written back every step, the injected quantization error is not a one-time perturbation; it can be recursively fed into later EMA updates.
We formulate this storage-level error accumulation mechanism below.
The goal is to isolate one instability mechanism in low-bit optimizer-state storage, not to provide a full convergence proof of the optimizer.

\subsection{Quantized-EMA Error Model}
\label{sec:ema_feedback}

Consider a full-precision EMA state and its low-bit stored counterpart:
\begin{equation}
\begin{aligned}
s_{t+1} &= \beta s_t + (1-\beta)u_{t+1}, \qquad 0<\beta<1,\\
s'_{t+1} &= \beta s_t^{dq} + (1-\beta)u_{t+1},\\
\epsilon_{t+1} &:= s_{t+1}^{dq}-s'_{t+1}, \qquad
e_t := s_t^{dq}-s_t .
\end{aligned}
\label{eq:ema_error_defs}
\end{equation}
Here $u_{t+1}$ is the per-step update term, $s_t^{dq}$ is the dequantized stored state, $\epsilon_{t+1}$ is the error injected when the updated state is quantized and written back, and $e_t$ is the accumulated state error relative to the full-precision EMA.
A direct subtraction gives
\begin{equation}
e_{t+1} = \beta e_t + \epsilon_{t+1}.
\label{eq:ema_error_recursion}
\end{equation}
Thus, injected errors are fed back through the EMA state rather than appearing as independent one-step perturbations.

\subsection{Theoretical Analysis: Mean and Noise Amplification in EMA}
\label{sec:ema_theory}

We now characterize how the EMA recursion propagates the bias and variance of the injected error.

\begin{lemma}[Bias induces persistent error under EMA recursion]
\label{lem:ema_mean}
Assume the injected errors satisfy $\mathbb{E}[\epsilon_t]=b$.
For the recursion $e_{t+1}=\beta e_t+\epsilon_{t+1}$ with $0<\beta<1$, we have
\begin{equation}
\mathbb{E}[e_t]
=
\beta^t\mathbb{E}[e_0]
+
\frac{1-\beta^t}{1-\beta}b
\;\;\longrightarrow\;\;
\frac{b}{1-\beta}.
\label{eq:ema_mean}
\end{equation}
In particular, even a small nonzero bias $b$ yields a persistent error whose magnitude scales as $O(\|b\|/(1-\beta))$ when $\beta\rightarrow 1$.
\end{lemma}

\begin{theorem}[Steady-state second moment under quantized EMA]
\label{thm:ema_second_moment}
Assume $\epsilon_t=b+\xi_t$, where $\{\xi_t\}_{t\geq 1}$ are independent with $\mathbb{E}[\xi_t]=0$ and $\mathbb{E}\|\xi_t\|_F^2=\sigma^2$.
Assume also that $e_0$ has finite second moment and is independent of $\{\xi_t\}_{t\geq 1}$.
For $0<\beta<1$, the state error satisfies
\begin{equation}
\lim_{t\rightarrow\infty}\mathbb{E}\|e_t\|_F^2
=
\underbrace{\frac{\|b\|_F^2}{(1-\beta)^2}}_{\text{mean (bias) term}}
+
\underbrace{\frac{\sigma^2}{1-\beta^2}}_{\text{variance (noise) term}} .
\label{eq:ema_second_moment}
\end{equation}
\end{theorem}

\begin{remark}[Mean dominates when $\beta\rightarrow 1$]
\label{rem:ema_mean_dominates}
The ratio between the mean term and the variance term in Eq.~\eqref{eq:ema_second_moment} equals
\begin{equation}
\frac{\|b\|_F^2/(1-\beta)^2}{\sigma^2/(1-\beta^2)}
=
\frac{(1+\beta)\|b\|_F^2}{(1-\beta)\sigma^2},
\label{eq:ema_mean_noise_ratio}
\end{equation}
which grows as $1/(1-\beta)$ when $\beta\approx 1$.
Thus, even a tiny nonzero mean in rounding error ($b\neq 0$) can dominate the long-horizon error budget for large-$\beta$ EMA states, such as AdamW's second moment.
\end{remark}

Proofs of Lemma~\ref{lem:ema_mean} and Theorem~\ref{thm:ema_second_moment} are given in Appendix~\ref{sec:proof}.
Eq.~\eqref{eq:ema_second_moment}--\eqref{eq:ema_mean_noise_ratio} motivates a two-step design rule for low-bit EMA states: first reduce systematic rounding bias, then control the variance.

\subsection{Real Setting: In-training Evidence}
\label{sec:real_setting}

\begin{figure}[t]
    \centering
    \begin{subfigure}[t]{0.49\linewidth}
        \centering
        \includegraphics[width=\linewidth,keepaspectratio]{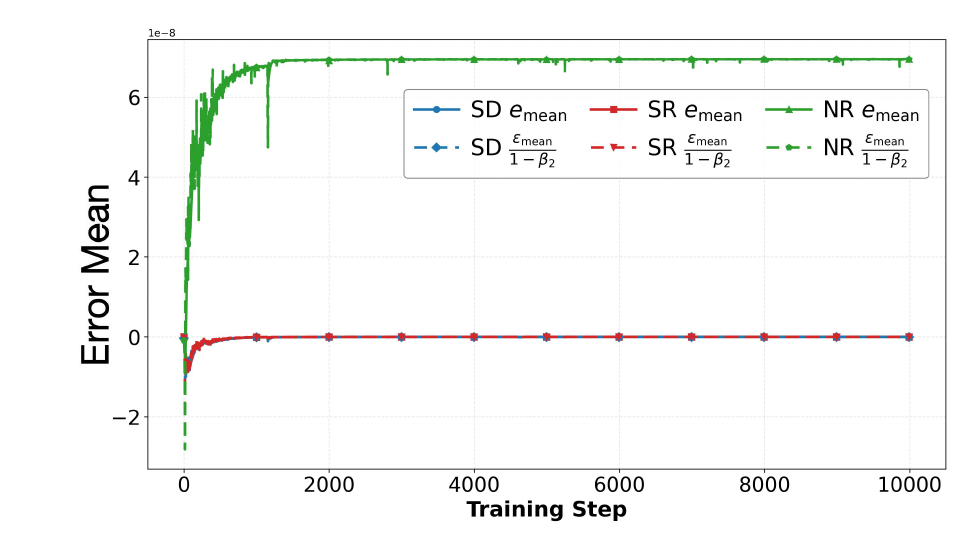}
        \caption{Mean drift matches $\mathbb{E}[\epsilon_t]/(1-\beta_2)$.}
        \label{fig:real_ema_mean}
    \end{subfigure}
    \hfill
    \begin{subfigure}[t]{0.49\linewidth}
        \centering
        \includegraphics[width=\linewidth,trim=0pt 0pt -20pt 0pt,clip,keepaspectratio]{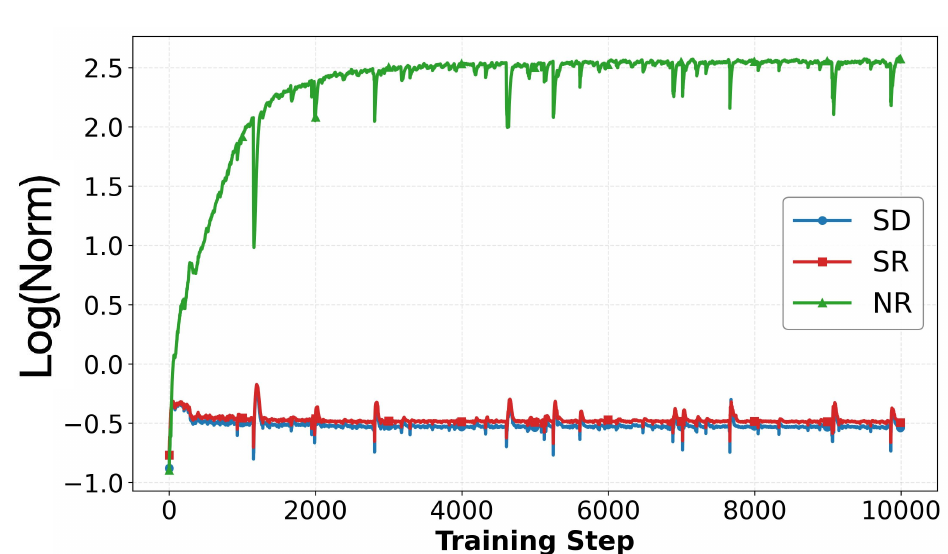}
        \caption{Relative error $\log_{10}(\|e_t\|_F/\|s_t\|_F)$.}
        \label{fig:real_ema_norm}
    \end{subfigure}
    \caption{
    In-training diagnostics for the quantized-EMA error model.
    During LLaMA-120M AdamW pretraining, we track the accumulated state error $e_t$ and injected write-back error $\epsilon_t$ for the second-moment state.
    (a) The measured mean error follows $\mathbb{E}[\epsilon_t]/(1-\beta_2)$ after the transient, consistent with EMA amplification of rounding bias.
    (b) NR accumulates large relative error, whereas SR and SD maintain bounded relative errors in this diagnostic, with SD slightly smaller.
    }
    \label{fig:real_ema_diagnostics}
\end{figure}

We verify that the same mechanism appears in real LLM training.
During LLaMA-120M pretraining with AdamW, we focus on the second-moment tensor $v$ in layer \texttt{11.attn.proj} and maintain measurement-only quantized EMA copies under NR, SR, and SD. These copies are recursively updated and requantized, but do not affect model updates or subsequent gradients. 
Using the definitions above, Eq.~\eqref{eq:ema_error_recursion} predicts the steady-state relation $\mathbb{E}[e_\infty]\approx \mathbb{E}[\epsilon_t]/(1-\beta_2)$ after the initial transient.
As shown in Fig.~\ref{fig:real_ema_diagnostics}, the measured $\mathbb{E}[e_t]$ closely tracks this prediction: NR exhibits persistent drift, while SR and SD keep the mean near zero.
The relative-error plot further shows that NR accumulates larger state error, whereas SR and SD maintain bounded relative errors in this diagnostic, with SD slightly smaller than SR.
Additional layer diagnostics are provided in Appendix~\ref{sec:fig_plot}.

\section{Method: BReD for EMA Optimizer States}
\label{sec:method}
\subsection{Overview}
Section \ref{sec:ema_error} shows that, when an EMA state is quantized and written back every step, the mean component of the injected error can be amplified by the EMA recursion, and the variance is also amplified. This motivates a simple design rule: remove systematic rounding bias first, then control the remaining variance.

We implement this rule as BReD, a rounding-centric framework built from three components: subtractive dithering (SD), deterministic seed replay (DSR), and block-wise randomization (BWR). SD gives unbiased reconstruction with controlled variance under the local non-clipped interval model. DSR deterministically reproduces the same dither value at dequantization time without storing random tensors. BWR amortizes random-number generation by sharing one random scalar within each block. 
Combining these components gives a practical subtractive-dithering module for EMA optimizer states. 
The same scheme can be applied to different EMA optimizer states by dequantizing for computation and requantizing for storage each step, without changing optimizer updates or hyperparameters.

\subsection{Element-wise View: Interval Parameterization}

Given a scale and a quantization grid, and ignoring clipping, the local quantization error is determined by the rounding and reconstruction rule. We therefore adopt an element-wise perspective on rounding a scalar to a quantization grid. Let $p_0(x)\leq x \leq p_1(x)$ be the two neighboring grid points that bracket $x$, and define the interval width and coordinate
\begin{equation}
\Delta(x) := p_1(x)-p_0(x), \qquad
\alpha(x) := \frac{x-p_0(x)}{\Delta(x)} \in [0,1].
\end{equation}
Hence, rounding can be written as choosing a binary endpoint indicator $c\in\{0,1\}$, followed by reconstruction of $\hat{\alpha}$ and $\hat{x}$:
\begin{equation}
c\in\{0,1\}, \qquad \hat{x}=p_0(x)+\Delta(x)\hat{\alpha}.
\end{equation}
The rounding--reconstruction gap is
\begin{equation}
\label{eq:error_gap}
\epsilon_{\rm elem}(x) := \hat{x}-x = \Delta(x)\big(\hat{\alpha}-\alpha(x)\big).
\end{equation}
A rounding rule is locally unbiased when $\mathbb{E}[\hat{\alpha}\mid \alpha]=\alpha$, which gives $\mathbb{E}[\epsilon_{\rm elem}\mid x]=0$ by Eq.(\ref{eq:error_gap}). Clipping or saturation can introduce additional error; in our comparisons, the scale, grid, and clipping behavior are kept identical across NR, SR, and BReD.

\subsection{Subtractive Dithering as an Unbiased then Low Variance Primitive}
\label{sec:round}
Section \ref{sec:ema_error} shows that EMA error feedback amplifies the mean component of quantization error. We therefore first enforce unbiased rounding, then reduce the remaining variance.

\paragraph{Nearest rounding (NR).}
NR deterministically selects the nearest endpoint:
\begin{equation}
c_{\rm nr}(\alpha) := \mathbbm{1}\{\alpha \geq 1/2\}, \qquad \hat{\alpha}=c_{\rm nr}(\alpha).
\end{equation}
NR is computationally efficient, but it can introduce biased rounding errors for asymmetric data distributions.

\paragraph{Stochastic rounding (SR).}
SR draws $u\sim \mathrm{Unif}(0,1)$ and selects the upper grid point with probability $\alpha$:
\begin{equation}
c_{\rm sr}(\alpha;u) := \mathbbm{1}\{u < \alpha\}, \qquad \hat{\alpha}=c_{\rm sr}(\alpha;u).
\end{equation}
Then $\mathbb{E}[\hat{\alpha}\mid \alpha]=\alpha$, hence $\mathbb{E}[\epsilon_{\rm elem}\mid x]=0$, and
\begin{equation}
\mathrm{Var}(\epsilon_{\rm elem}\mid x)=\Delta(x)^2\alpha(1-\alpha)\leq \Delta(x)^2/4.
\end{equation}
Although SR achieves unbiasedness, it can still leave substantial residual noise, which is also accumulated by the EMA recursion.

\paragraph{Subtractive dithering (SD).}
SD draws $u\sim \mathrm{Unif}(0,1)$, rounds a dithered coordinate to a grid point, and reconstructs a continuous value using the same $u$:
\begin{equation}
\label{eq:sd_sample}
c_{\rm sd}(\alpha;u) := \mathbbm{1}\{\alpha+u\geq 1\}\in\{0,1\},
\end{equation}
\begin{equation}
\label{eq:sd_round}
\hat{\alpha}:=c_{\rm sd}(\alpha;u)-u+\frac{1}{2},
\qquad
\hat{x}:=p_0(x)+\Delta(x)\hat{\alpha}.
\end{equation}
SD is unbiased, i.e., $\mathbb{E}[\epsilon_{\rm elem}\mid x]=0$, and has
\begin{equation}
\mathrm{Var}(\epsilon_{\rm elem}\mid x)=\Delta(x)^2/12,
\end{equation}
which is independent of $\alpha$. Therefore, SD does not dominate SR pointwise in conditional variance, but it provides $\alpha$-independent noise and a smaller worst-case conditional variance ($\Delta(x)^2/12$ versus $\Delta(x)^2/4$ for SR). A short proof is deferred to Appendix \ref{proof:sd}.

Under the local non-clipped interval model, SR errors can approach
one bin width, whereas SD errors are bounded by half a bin
(Appendix~\ref{app:rounding_tails}).
This difference offers a possible explanation for the observed
SR failures, but is not a divergence criterion derived from
Theorem~\ref{thm:ema_second_moment}.

\subsection{Deterministic Seed Replay (DSR): Stateless Reconstruction of Dither Values}

Subtractive dithering requires using the same random dither value $u$ at quantization and dequantization; otherwise the subtractive correction does not cancel. Storing $u$ per element would introduce an extra tensor comparable to the state itself, defeating compression. We therefore regenerate dither values on the fly via a stateless keyed pseudo-random number generator (PRNG).

For each EMA state tensor, we assign a fixed identifier \texttt{state\_id}, such as a tuple containing the optimizer-state type and tensor index. For AdamW, for example, the state type distinguishes the first and second moments; the tensor index distinguishes different parameter tensors. When the state is written at training step $t$, for each block index $j$, we form a deterministic key
\begin{equation}
\kappa(j,t) := (\texttt{global\_seed}, \texttt{state\_id}, t, j),
\end{equation}
and draw the dither as
\begin{equation}
\label{eq:prng}
u \leftarrow \mathrm{PRNG}(\kappa(j,t)) \in (0,1).
\end{equation}
When decoding the stored state written at step $t$, we reuse the same key $\kappa(j,t)$ to replay the identical $u$, thus enabling subtractive dithering without storing random tensors.

\subsection{Block-wise Randomization (BWR): Amortizing Randomness}

Randomized rounding, including SR and SD, requires generating random numbers at scale. To reduce random-number generation overhead, BWR shares one random scalar within each block.

Let $\{B_k\}_{k=1}^K$ be a partition of indices into blocks of size $B$. Instead of sampling i.i.d. randomness per coordinate, BWR samples one $u^{(k)}\sim \mathrm{Unif}(0,1)$ per block and reuses it for all $i\in B_k$. For example, block-wise SR uses
\begin{equation}
c_i := \mathbbm{1}\{u^{(k)} < \alpha(x_i)\}, \qquad i\in B_k.
\end{equation}
For SD, the same block dither $u^{(k)}$ is used in Eq.\ref{eq:sd_sample}--\ref{eq:sd_round} for all elements in the block and is replayed during dequantization. If the underlying element-wise rule is unbiased, BWR preserves coordinate-wise marginal unbiasedness after taking expectation over $u^{(k)}$. BWR reduces PRNG calls from $O(d)$ to $O(d/B)$ while introducing within-block correlation, whose impact we study via block-size ablations.

\subsection{BReD: Block Replay Dithering (SD + DSR + BWR)}

We now introduce \textbf{BReD} (\textbf{B}lock \textbf{Re}play \textbf{D}ithering), a practical method for EMA optimizer-state quantization. BReD combines: (i) SD to achieve unbiased reconstruction error with controlled variance; (ii) DSR to replay the same dither values at dequantization time without storing random tensors; and (iii) BWR to amortize PRNG calls by sharing one dither value per block.

At write time, for each block written at training step $t$, BReD generates a single dither value via the keyed PRNG in Eq.\ref{eq:prng}, applies subtractive dithering to the elements in the block, and stores the resulting low-bit codes together with the usual block scale. At read time, the same $u^{(k)}$ is replayed to decode the stored values and reconstruct the state. BReD can be combined with other scaling methods; in this work, we primarily use Microscaling data formats \citep{rouhani2023ocp, rouhani2023microscaling}.

\section{Experiments}

We evaluate BReD for low-bit storage of EMA optimizer states in pretraining and supervised fine-tuning. 
The pretraining experiments follow four questions: whether the rounding rule matters under a fixed scaling rule and quantization grid; whether BReD can be combined with optimizer-specific low-bit methods; whether stability persists at larger scales and longer token horizons; and how BReD changes the memory--perplexity trade-off. 
We then evaluate supervised fine-tuning and ablate the block-wise randomization design.

Unless otherwise stated, all low-bit variants use the same optimizer update and hyperparameters as the corresponding full-precision optimizer-state baseline. 
The only change is how the selected EMA optimizer states are stored between training steps.
Standard pretraining and fine-tuning results report
mean $\pm$ standard deviation over three runs.
The 100B-token experiment uses one run due to compute cost.
The additional high-$\beta$ check uses one paired run,
and each SOLO and Shampoo configuration is evaluated once.

\subsection{Experimental setup}

\paragraph{Training setups.}
For pretraining, we evaluate AdamW \citep{kingma2014adam,loshchilov2017decoupled}, Muon \citep{jordan6muon}, NAdamW \citep{dozat2016incorporating}, SOAP \citep{vyas2024soap}, and MARS \citep{yuan2024mars} at the 120M scale, and AdamW and Muon at the 1.1B and 3.4B scales. 
The 120M and 1.1B models use TinyLlama-style configurations, while the 3.4B model follows OpenLLaMA. 
Models are trained on SlimPajama for 10.5B tokens at 120M and 31.4B tokens at 1.1B/3.4B. 
We report validation perplexity (PPL) and zero-shot accuracies on HellaSwag~\citep{zellers2019hellaswag}, OpenBookQA~\citep{mihaylov2018can}, WinoGrande~\citep{sakaguchi2021winogrande}, ARC~\citep{Clark2018ThinkYH}, BoolQ~\citep{clark2019boolq}, and PIQA~\citep{bisk2020piqa}.
For supervised fine-tuning, we fine-tune Qwen2.5-7B~\citep{qwen2025qwen25technicalreport} on UltraFeedback~\citep{cui2023ultrafeedback} and evaluate HumanEval~\citep{chen2021evaluating}, MBPP~\citep{austin2021program}, GSM8K~\citep{cobbe2021training}, TruthfulQA \citep{lin2022truthfulqa}, and validation PPL. Our fine-tuning recipe follows prior work on instruction-tuning~\citep{yu2023metamath, liu2023makes, cui2023ultrafeedback}. 
% All hyperparameters are provided in Appendix \ref{sec:sft-hyper}, task-wise results are provided in Appendix~\ref{sec:exp_tab}.
Optimizer hyperparameters are provided in
Appendices~\ref{app:optim_hparams} and~\ref{sec:sft-hyper};
task-wise pretraining results are provided in
Appendix~\ref{sec:exp_tab}.

\paragraph{Quantized states and baselines.}
BReD quantizes EMA optimizer states, such as the first and second moments in AdamW and the momentum states in Muon-style optimizers. 
For SOAP, we additionally quantize the preconditioner states. 
Following prior low-bit optimizer-state work \citep{dettmers20218}, we keep embedding-layer optimizer states in full precision during pretraining. 
Unless otherwise stated, NR, SR, and BReD use the same block scaling rule and quantization grid. 
For 4-bit storage, we use MXFP4 by default; for 3-bit storage,
we use the grid $\{0,\pm1,\pm2,\pm3\}$ with Microscaling.
An additional MXINT4 check with 120M AdamW also converges,
reaching a validation PPL of 18.9
(Appendix~\ref{app:mxint4}).
We compare rounding baselines (NR/SR), optimizer-specific low-bit methods (AdamW-LPMM~\citep{li2023memory} and Muon-Grasp\citep{wuachieving}).

\subsection{Pretraining: rounding controls and low-bit methods}

We first isolate the rounding rule under a fixed scaling rule, quantization grid, optimizer update, and hyperparameters. 
As shown in Table~\ref{tab:pretrain-rounding-lowbit}(a), NR and SR fail or substantially degrade in all five 120M optimizer settings: four optimizers produce NaN or divergence, and Muon completes training only with much worse PPL. 
In contrast, BReD remains stable across all five optimizers, with validation PPL between 18.0 and 18.9. 
This controlled comparison shows that stability in aggressive 4-bit EMA-state storage is not a consequence of the low-bit format alone.
An additional paired run with AdamW at $\beta_2=0.999$
also remains stable, with a PPL increase of $0.7$ relative
to the matched FP32-state baseline
(Appendix~\ref{app:high_beta}).
An additional single-run check for each configuration with
120M Shampoo also finds that BReD completes training with
4-bit preconditioner states, whereas NR and SR diverge
(Appendix~\ref{app:shampoo}).

We next test whether BReD can be combined with optimizer-specific low-bit methods. 
Table~\ref{tab:pretrain-rounding-lowbit}(b) shows that adding BReD improves AdamW-LPMM from 18.7 to 18.4 PPL, and matches the best observed Muon-Grasp PPL of 17.8 while increasing average zero-shot accuracy from 40.6 to 41.2. 
Thus, BReD can serve as a rounding component for EMA-state write-back rather than only as a standalone MXFP4 state wrapper.
We also evaluate SOLO~\citep{xu2025pushing} in our 120M
pretraining setup (Appendix~\ref{app:solo}).
We report these checks separately from the controlled rounding
comparison because SOLO also changes the quantization grids
and first-moment dynamics.

\begin{table}[t]
\centering
\small
\caption{
Pretraining controls at the 120M scale. 
(a) Controlled 4-bit rounding comparison. NR, SR, and BReD use the same scaling rule, quantization grid, optimizer update, and hyperparameters. 
Numbers in parentheses are validation PPL. 
(b) Combination with optimizer-specific low-bit methods. 
}
\label{tab:pretrain-rounding-lowbit}

\begin{tabular}{l|c|ccc}
\toprule
\multicolumn{4}{l}{\textit{(a) Controlled 4-bit rounding comparison}} \\
\midrule
Optimizer & FP32-State &NR & SR & BReD \\
\midrule
AdamW  & 18.3 &Diverged          & Diverged          & \textbf{Stable (18.7)} \\
Muon   & 17.8 &Degraded (31.0) & Degraded (21.7) & \textbf{Stable (18.0)} \\
NAdamW & 18.9 &Diverged          & Diverged          & \textbf{Stable (18.9)} \\
SOAP   & 17.6 &Diverged          & Diverged          & \textbf{Stable (18.1)} \\
MARS   & 18.7 &Diverged          & Diverged          & \textbf{Stable (18.9)} \\
\bottomrule
\end{tabular}

\vspace{0.5em}

\begin{tabular}{lcccc}
\toprule
\multicolumn{5}{l}{\textit{(b) Combination with optimizer-specific low-bit methods}} \\
\midrule
\multirow{2}{*}{Method} 
& \multicolumn{2}{c}{Avg. acc. $\uparrow$} 
& \multicolumn{2}{c}{PPL $\downarrow$} \\
\cmidrule(lr){2-3} \cmidrule(lr){4-5}
& Base & +BReD & Base & +BReD \\
\midrule
AdamW-LPMM & 40.3 & 40.5 (+0.2) & 18.7 & 18.4 (-0.3) \\
Muon-Grasp & 40.6 & 41.2 (+0.6) & 17.8 & 17.8 (+0.0) \\
\bottomrule
\end{tabular}
\end{table}

\subsection{Pretraining: scale, horizon, and memory trade-offs}

We next test whether 4-bit BReD remains stable beyond the 120M pretraining runs. 
Table~\ref{tab:scale-check} reports larger-scale results for AdamW and Muon at 1.1B and 3.4B. 
Across these settings, BReD remains close to the corresponding full-precision optimizer-state baselines, with PPL shifts no larger than 0.3 and average zero-shot accuracy changes within 0.5 points. Details can be found in Table \ref{tab:1b_3b_result}.

\begin{figure}[t]
\centering
\begin{minipage}[t]{0.43\linewidth}
\centering
\includegraphics[
    width=\linewidth,
    trim=18 17 18 10,
    clip
]{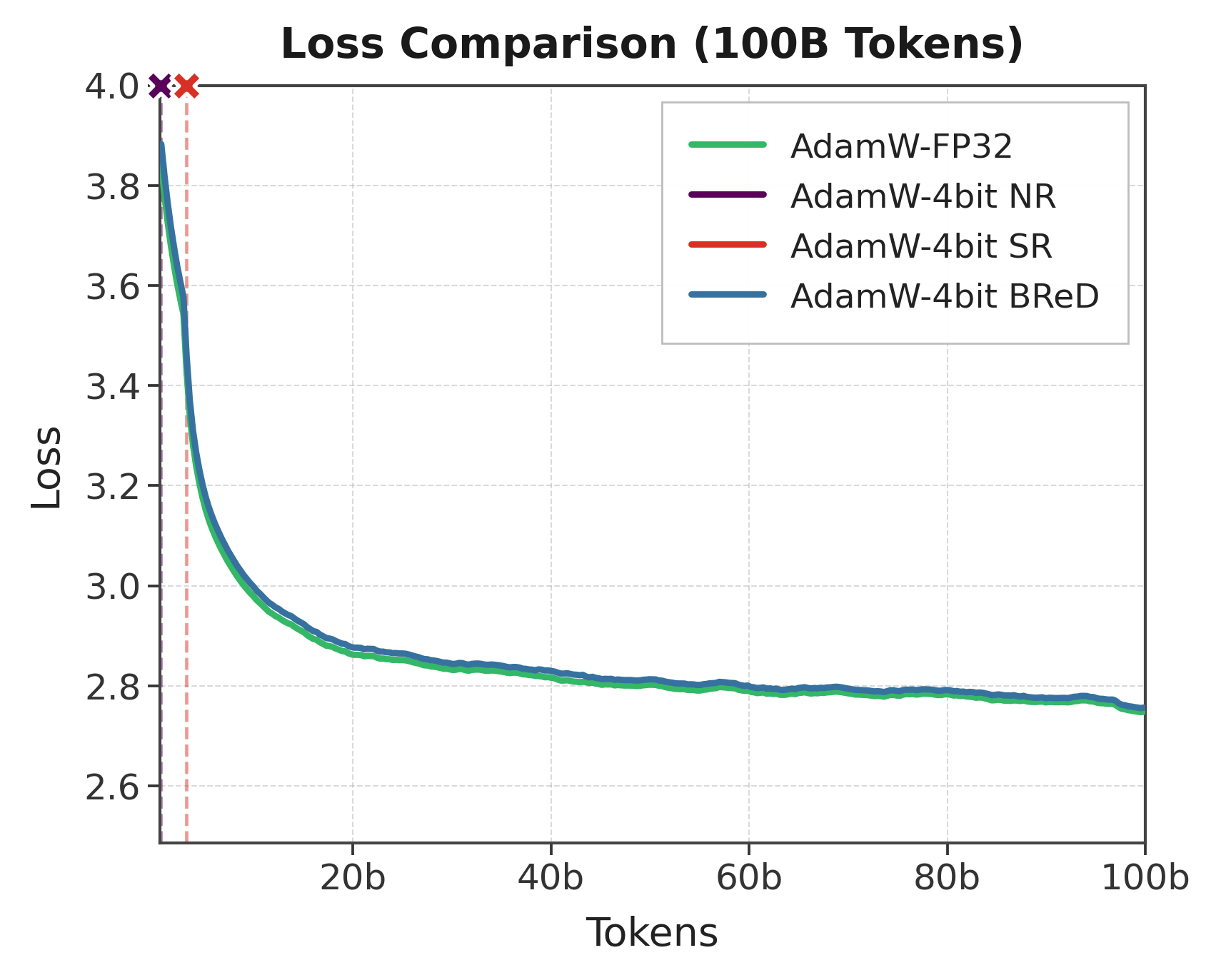}
\vspace{-0.em}
\centerline{\small (a) 100B-token AdamW loss}
\end{minipage}
\hfill
\begin{minipage}[t]{0.49\linewidth}
\centering
\includegraphics[
    width=\linewidth,
    clip
]{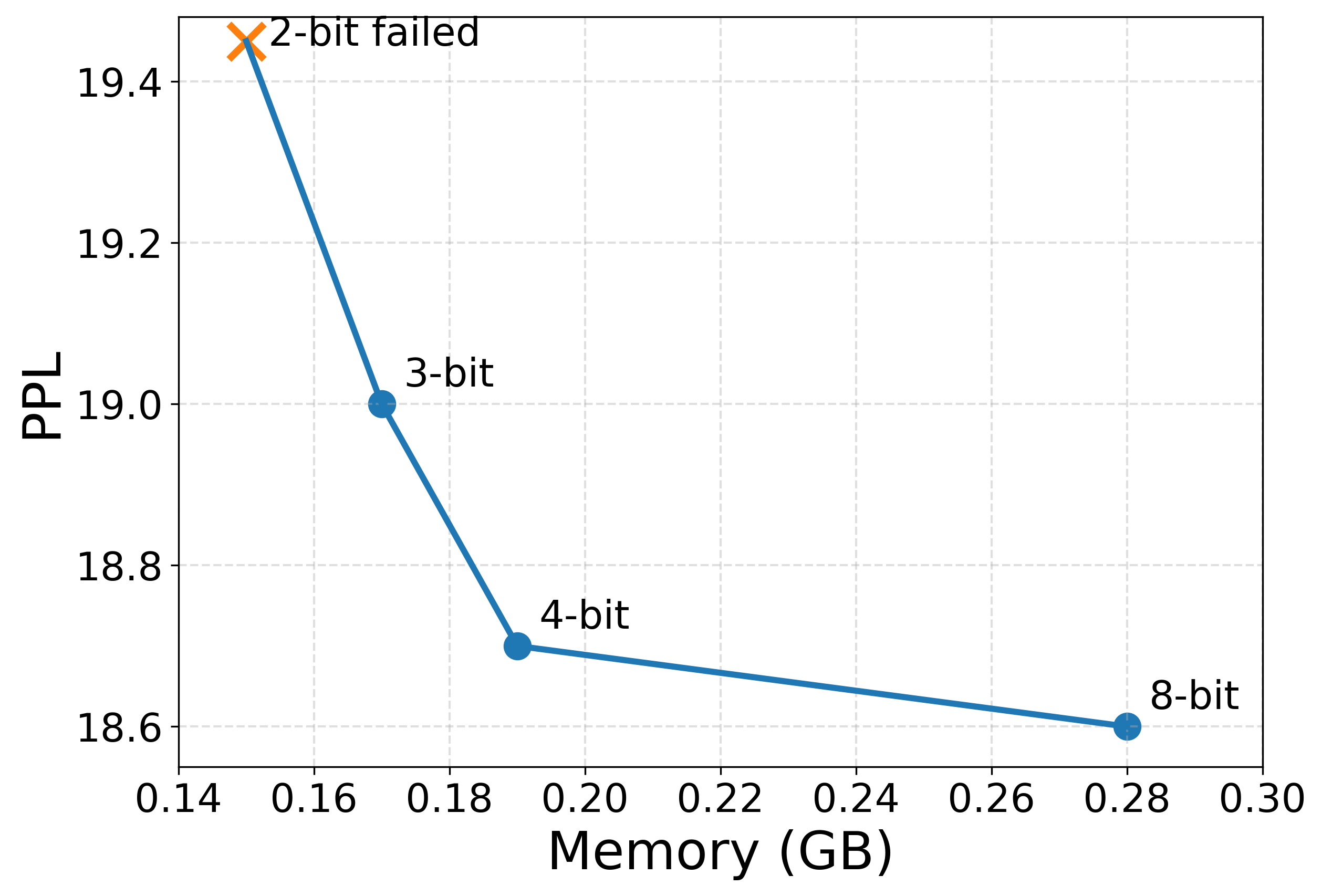}
\vspace{-0.em}
\centerline{\small (b) 120M AdamW Pareto}
\end{minipage}
\vspace{-0.4em}
\caption{
Pretraining horizon and memory-quality checks. 
(a) Longer-horizon AdamW pretraining at the 120M scale. 
(b) Memory-perplexity trade-off for 120M AdamW pretraining. 
The Pareto plot is a case study; broader stability evidence is provided by the controlled and scale experiments.
}
\label{fig:horizon-pareto}
\end{figure}

We also run a longer-horizon AdamW pretraining test for 100B tokens at the 120M scale. 
As shown in Figure~\ref{fig:horizon-pareto}(a), 4-bit BReD remains close to the full-precision optimizer-state loss curve throughout training. 
This experiment is intentionally scoped to AdamW, but uses a much longer token horizon than the standard 120M runs.

Finally, we report a memory-perplexity case study for 120M AdamW in Figure~\ref{fig:horizon-pareto}(b). 
Moving from the 8-bit state-storage setting to 4-bit BReD reduces optimizer-state memory from 0.28GB to 0.19GB while keeping validation PPL close.
% 4-bit BReD reduces optimizer-state memory from 0.28GB to 0.19GB while keeping validation PPL close to the 8-bit setting. 
Moving to 3-bit further reduces memory to 0.17GB but increases validation PPL to 19.0, while the 2-bit setting fails. 
This 2-bit failure is specific to the evaluated recipe and
does not establish a fundamental lower bound for subtractive
dithering.
Across the evaluated optimizers and scales, packed 4-bit BReD reduces optimizer-state memory by 35.9\%--86.0\%; the full memory table is provided in Appendix~\ref{sec:memory}.

\begin{table}[t]
\centering
\small
\caption{
Scale check for 4-bit BReD in pretraining. 
Parentheses denote changes relative to the corresponding full-precision optimizer-state baseline.
}
\label{tab:scale-check}
\begin{tabular}{llcccc}
\toprule
\multirow{2}{*}{Scale} 
& \multirow{2}{*}{Optimizer}
& \multicolumn{2}{c}{Avg. acc. $\uparrow$}
& \multicolumn{2}{c}{PPL $\downarrow$} \\
\cmidrule(lr){3-4} \cmidrule(lr){5-6}
& & FP32-state & 4-bit BReD & FP32-state & 4-bit BReD \\
\midrule
1.1B & AdamW & 46.5 & 46.4 (-0.1) & 10.8 & 10.9 (+0.1) \\
1.1B & Muon  & 47.6 & 47.1 (-0.5) & 10.3 & 10.6 (+0.3) \\
\midrule
3.4B & AdamW & 47.9 & 47.6 (-0.3) & 9.7 & 9.8 (+0.1) \\
3.4B & Muon  & 48.2 & 47.7 (-0.5) & 9.6 & 9.8 (+0.2) \\
\bottomrule
\end{tabular}
\end{table}

\paragraph{End-to-end efficiency.}
Using the default training configurations reported in the paper, Table~\ref{tab:runtime-memory} reports matched FP32-state and
4-bit BReD measurements for AdamW.
BReD reduces persistent optimizer-state storage by
78.9\%--86.0\%, with throughput reductions of 5.1\%--16.0\%.
Timing includes forward, backward, and all optimizer-state
processing, including decoding and requantization, but excludes
compilation, evaluation, and checkpointing.
For gradient accumulation, micro-steps are aggregated into
optimizer steps.
The relative overhead is smaller in the larger configurations,
but model size and gradient accumulation change together,
so this trend cannot be attributed solely to model scale.

\begin{table}[t]
\centering
\small
\setlength{\tabcolsep}{4pt}
\caption{
End-to-end AdamW efficiency.
FP32 and BReD denote full-precision and 4-bit optimizer-state
storage, respectively.
Memory reports persistent optimizer-state storage,
not peak device memory.
}
\label{tab:runtime-memory}
\begin{tabular}{lrrrrrr}
\toprule
& \multicolumn{2}{c}{Step time (s)}
& \multicolumn{2}{c}{Throughput (M tok/s)}
& \multicolumn{2}{c}{State memory (GB)} \\
\cmidrule(lr){2-3}
\cmidrule(lr){4-5}
\cmidrule(lr){6-7}
Scale & FP32 & BReD & FP32 & BReD & FP32 & BReD \\
\midrule
120M & 0.547  & 0.651  & 1.917 & 1.611 & 0.90  & 0.19 \\
1.1B & 2.881  & 3.178  & 0.364 & 0.330 & 8.20  & 1.18 \\
3.4B & 10.386 & 10.945 & 0.101 & 0.096 & 25.53 & 3.58 \\
\bottomrule
\end{tabular}
\end{table}

\subsection{Supervised fine-tuning}

We next test whether the same optimizer-state storage rule transfers beyond pretraining. 
We fine-tune Qwen2.5-7B on UltraFeedback and evaluate HumanEval, MBPP, GSM8K, TruthfulQA, and validation PPL. 
As shown in Table~\ref{tab:sft}, 4-bit BReD remains close to the corresponding full-precision optimizer-state baselines for both AdamW and Muon. 
For AdamW, the average downstream score changes from 54.2 to 54.9 with no change in validation PPL. 
For Muon, the average downstream score changes from 55.2 to 54.9, again with no change in validation PPL. 
These results suggest that BReD can be applied in a downstream fine-tuning recipe without changing optimizer hyperparameters.

\begin{table}[t]
\centering
\small
\caption{
Supervised fine-tuning with 4-bit BReD-quantized optimizer states. 
We fine-tune Qwen2.5-7B on UltraFeedback and report average downstream score and validation PPL. 
Parentheses denote changes relative to the corresponding full-precision optimizer-state baseline.
}
\label{tab:sft}
\begin{tabular}{lcccc}
\toprule
\multirow{2}{*}{Optimizer}
& \multicolumn{2}{c}{Avg. score $\uparrow$}
& \multicolumn{2}{c}{PPL $\downarrow$} \\
\cmidrule(lr){2-3} \cmidrule(lr){4-5}
& FP32-state & 4-bit BReD & FP32-state & 4-bit BReD \\
\midrule
AdamW & 54.2 & 54.9 (+0.7) & 4.6 & 4.6 (+0.0) \\
Muon  & 55.2 & 54.9 (-0.3) & 4.6 & 4.6 (+0.0) \\
\bottomrule
\end{tabular}
\end{table}

\subsection{Aggressive 3-bit compression and ablations}

The main results use 4-bit BReD as the default operating point. 
We also evaluate 3-bit BReD as a more aggressive compression setting. 
As summarized in Table~\ref{tab:threebit-ablation}(a), 3-bit BReD remains convergent in all evaluated pretraining and fine-tuning settings, but the quality shifts are larger and less uniform than in the 4-bit case. 
We therefore treat 3-bit BReD as an aggressive memory-quality trade-off rather than the default setting; results for each evaluated setting are provided in
Appendix~\ref{sec:apd_3bit_results}.

We further ablate the block-wise randomization component of BReD. 
BWR shares one dither value within each block, reducing PRNG calls from $O(d)$ to $O(d/B)$ while introducing within-block correlation. 
Table~\ref{tab:threebit-ablation}(b) shows that the default $B=32$ remains close to element-wise dithering in quality, while reducing the number of generated random values by $32\times$ and lowering generation time from $486.32{\pm}29$ ms to $340.88{\pm}41$ ms.

\begin{table}[t]
\centering
\small
\caption{
Aggressive compression and BWR ablation. 
(a) Summary of 3-bit BReD results. 
$\Delta$Avg. denotes the range of changes relative to the corresponding full-precision optimizer-state baseline. 
(b) Block-wise randomization ablation for 4-bit AdamW at the 120M scale. 
}
\label{tab:threebit-ablation}

\begin{tabular}{llccc}
\toprule
\multicolumn{5}{l}{\textit{(a) 3-bit BReD summary}} \\
\midrule
Setting & Architecture / size & Optimizers & $\Delta$Avg. & $\Delta$PPL \\
\midrule
Pretrain & LLaMA-style / 120M & 5 opt. & $[+0.0,+1.0]$ & $[+0.1,+1.2]$ \\
Pretrain & LLaMA-style / 1.1B & AdamW, Muon & $[-0.9,-0.7]$ & $[+0.1,+0.6]$ \\
SFT & Qwen2.5 / 7B & AdamW, Muon & $[-0.8,+0.8]$ & $[-0.1,+0.1]$ \\
\bottomrule
\end{tabular}

\vspace{0.6em}

\begin{tabular}{lcccc}
\toprule
\multicolumn{5}{l}{\textit{(b) BWR ablation}} \\
\midrule
Block size $B$ & Avg. acc. $\uparrow$ & PPL $\downarrow$ & Random ratio & Generation time \\
\midrule
1   & 40.9 & 18.8 & $1$     & $486.32{\pm}29$ ms \\
32  & 40.7 & 18.7 & $1/32$  & $340.88{\pm}41$ ms \\
256 & 40.6 & 18.8 & $1/256$ & $335.47{\pm}35$ ms \\
\bottomrule
\end{tabular}
\end{table}

\section{Conclusion}
We studied a storage-level failure mode in low-bit EMA optimizer states: stepwise quantization and write-back can accumulate biased rounding error and amplify residual noise when the EMA coefficient is large.
We proposed \textbf{BReD} (\textbf{B}lock \textbf{Re}play \textbf{D}ithering), which makes subtractive dithering practical with deterministic seed replay and block-wise shared dither values, without changing optimizer updates.
Across pretraining and fine-tuning, BReD provides robust 4-bit optimizer-state storage and stable 3-bit convergence in the evaluated settings.

\section{Limitations}
% While the results support BReD as a practical low-bit quantization method for EMA states, there are limitations remain.
Due to limited computational resources, 
our largest pretraining experiments use a 3.4B model
and cover only AdamW and Muon.
% our strongest evidence is on 3.4B model and a selected set of optimizers, 
Broader validation at larger scales (e.g., tens of billions of parameters), longer token budgets, and additional optimizer families would further strengthen the conclusions.

Cross-optimizer breadth is evaluated at 120M, whereas larger-scale
pretraining covers only AdamW and Muon.
We do not provide matched comparisons with Adafactor~\citep{shazeer2018adafactor}, Adam-mini~\citep{zhang2024adam},
4-bit Shampoo~\citep{wang20244}, or
4-bit Preconditioned Optimization~\citep{li2025memory}.
Joint use with gradient or activation quantization remains
unevaluated.

Our storage-error analysis does not characterize how quantization
errors affect eigendecomposition or other spectral operations
on preconditioner states.

\section*{Acknowledgment}
Heshen Zhan would like to thank Qihao Cheng and Xuefei Wang for the discussions of experimental observations.

This paper is supported by NSFC (No. 12326608); Hetao Shenzhen-Hong Kong Science and Technology Innovation Cooperation Zone Project (No.HZQSWS-KCCYB-2024016); Guangdong Provincial Key Laboratory of Mathematical Foundations for Artificial Intelligence (2023B1212010001).

\bibliography{example_paper}
\bibliographystyle{plainnat}

%%%%%%%%%%%%%%%%%%%%%%%%%%%%%%%%%%%%%%%%%%%%%%%%%%%%%%%%%%%%

\newpage
\appendix

\section{Pre-training Optimizer Hyperparameters}
\label{app:optim_hparams}

This appendix reports the complete optimizer hyperparameters used in all experiments.
Unless otherwise stated, we keep \textbf{all training settings identical} across methods
(model architecture, data recipe, training length, batch size, learning-rate schedule, and evaluation),
and only vary the \textbf{storage precision} of optimizer states.
\subsection{Global settings}
\label{app:optim_global}
All runs share the following optimizer-related settings:
\begin{itemize}
  \item \textbf{Gradient clipping:} global norm clip at $1.0$ (disabled if set to $0$).
  \item \textbf{Epsilon:} $\epsilon = 1 \times 10^{-8}$.
  \item \textbf{LR scheduler:} WSD~\citep{hu2024minicpm,wen2024understanding}.
  \item \textbf{Warmup ratio:} 0.1 (warmup steps = $0.1 \times$ total steps).
  \item \textbf{Min LR ratio:} 0.1 (minimum LR = $0.1 \times$ peak LR).
  \item \textbf{Batch Size:} 1024.
  \item \textbf{Sequence length:} 1024.
\end{itemize}

\subsection{Hyperparameters for 120M pretraining}
Table~\ref{tab:120m_optim_hparams} summarizes the optimizer hyperparameters used for pretraining the 120M model.
Unless otherwise noted, all unspecified entries are not applicable to the corresponding optimizer.

\begin{table}[h]
  \centering
  \caption{Optimizer hyperparameters for the 120M pretraining runs.}
  \label{tab:120m_optim_hparams}
  \begin{tabular}{lccccc}
    \toprule
    & AdamW & Muon & NAdamW & SOAP & MARS \\
    \midrule
    Peak LR & $1\times10^{-3}$ & $1\times10^{-3}$ & $1\times10^{-3}$ & $2\times10^{-3}$ & $1\times10^{-3}$ \\
    $(\beta_1, \beta_2)$ & (0.9, 0.95) & \textemdash & (0.9, 0.95) & (0.95, 0.95) & (0.9, 0.99) \\
    Weight decay & 0.1 & 0.1 & 0.1 & 0.01 & 0 \\
    Weight decay (1D params) & \textemdash & \textemdash & \textemdash & \textemdash & 0.01 \\
    Muon momentum $\beta$ & \textemdash & 0.95 & \textemdash & \textemdash & \textemdash \\
    NS steps & \textemdash & 5 & \textemdash & \textemdash & \textemdash \\
    Momentum decay & \textemdash & \textemdash & 0.004 & \textemdash & \textemdash \\
    Shampoo $\beta$ & \textemdash & \textemdash & \textemdash & 0.95 & \textemdash \\
    \bottomrule
  \end{tabular}
\end{table}

\subsection{Settings for larger models}
Following \citet{liu2025muon}, we use the same learning rate and weight decay for Muon and AdamW.
For the 1.1B and 3.4B models, we use:
\begin{itemize}
  \item \textbf{Peak LR for 1.1B model:} $6\times 10^{-4}$.
  \item \textbf{Peak LR for 3.4B model:} $6\times 10^{-4}$.
  \item \textbf{AdamW betas:} $(\beta_1, \beta_2) = (0.9, 0.95)$.
  \item \textbf{Muon momentum:} $\beta = 0.95$.
  \item \textbf{NS steps:} 5.
\end{itemize}

\section{Fine-tuning hyperparameters}
\label{sec:sft-hyper}
Following \cite{yu2023metamath,liu2023makes,li2024preserving}, we use the learning rate $2*10^{-5}$ with a cosine learning rate decay schedule for 3 epochs. The warmup ratio is 0.03. The batch size is 128.

\subsection{Packed low-bit storage.}
All reported BReD low-bit runs use physically packed state tensors rather than high-precision emulation. 
We implement custom Triton kernels that store encoded optimizer states in packed integer tensors, decode them on the fly during the optimizer step, perform the EMA update in higher precision, and requantize and pack the updated states back to memory. 

\newpage
\section{Experiments results}
\label{sec:exp_tab}

\subsection{Task-wise pretraining results}

\begin{table}[h]
  \caption{
  Zero-shot LM evaluation accuracies for the 120M model.
  Subscripts denote standard deviations over three runs.
  Parentheses report the difference relative to the full-precision optimizer-state baseline of the same optimizer.
  }
  \label{tab:pretrain_result_120m}
  \setlength{\tabcolsep}{2.5pt}
  \begin{center}
    \begin{small}
      \begin{sc}
        \begin{tabular}{c|ccccccc|c}
          \toprule
          Optimizer  & HS & OBQA & WG  & Arc\_e & Arc\_c & BoolQ & PIQA  & Avg.$\uparrow$ \\
          \midrule
          AdamW    & 29.41$_{\textcolor{gray}{0.4}}$  & 27.60$_{\textcolor{gray}{0.7}}$ & 51.20$_{\textcolor{gray}{0.4}}$ & 36.36$_{\textcolor{gray}{0.1}}$ & 21.79$_{\textcolor{gray}{1.1}}$ & 56.12$_{\textcolor{gray}{1.9}}$ & 60.58$_{\textcolor{gray}{0.6}}$ & 40.4(+0.0) \\
          AdamW-4bit+\textbf{BReD} & 29.12$_{\textcolor{gray}{0.2}}$ & 27.07$_{\textcolor{gray}{0.6}}$ & 52.04$_{\textcolor{gray}{0.7}}$ & 35.84$_{\textcolor{gray}{1.0}}$ & 21.82$_{\textcolor{gray}{0.7}}$ & 59.35$_{\textcolor{gray}{1.0}}$ & 59.58$_{\textcolor{gray}{0.5}}$ & 40.7(+0.3) \\
          AdamW-3bit+\textbf{BReD} & 28.68$_{\textcolor{gray}{0.2}}$ & 27.27$_{\textcolor{gray}{0.5}}$ & 51.93$_{\textcolor{gray}{0.2}}$ & 36.10$_{\textcolor{gray}{0.9}}$ & 22.33$_{\textcolor{gray}{0.6}}$ & 59.63$_{\textcolor{gray}{1.3}}$ & 58.62$_{\textcolor{gray}{0.6}}$ & 40.6(+0.2)  \\
          \midrule
          Muon & 29.85$_{\textcolor{gray}{0.1}}$ & 26.73$_{\textcolor{gray}{0.4}}$ & 51.30$_{\textcolor{gray}{0.5}}$ & 36.60$_{\textcolor{gray}{0.3}}$ & 22.50$_{\textcolor{gray}{0.3}}$ & 59.16$_{\textcolor{gray}{2.6}}$ & 60.19$_{\textcolor{gray}{0.3}}$ & 40.9(+0.0) \\
          Muon-4bit+\textbf{BReD} & 29.44$_{\textcolor{gray}{0.2}}$ & 27.13$_{\textcolor{gray}{1.5}}$ & 52.57$_{\textcolor{gray}{1.1}}$ & 36.67$_{\textcolor{gray}{1.3}}$ & 22.55$_{\textcolor{gray}{0.3}}$ & 60.15$_{\textcolor{gray}{1.7}}$ & 60.08$_{\textcolor{gray}{0.9}}$ & 41.2(+0.3)  \\
          Muon-3bit+\textbf{BReD} & 29.02$_{\textcolor{gray}{0.1}}$ & 27.47$_{\textcolor{gray}{0.6}}$ & 51.49$_{\textcolor{gray}{0.8}}$ & 36.74$_{\textcolor{gray}{0.4}}$ & 22.24$_{\textcolor{gray}{0.5}}$ & 59.67$_{\textcolor{gray}{0.6}}$ & 59.99$_{\textcolor{gray}{0.4}}$ & 40.9(+0.0) \\
          \midrule
          NAdamW & 28.91$_{\textcolor{gray}{0.5}}$ & 26.47$_{\textcolor{gray}{0.4}}$ & 50.85$_{\textcolor{gray}{0.9}}$ & 36.08$_{\textcolor{gray}{0.4}}$ & 21.96$_{\textcolor{gray}{0.9}}$ & 56.71$_{\textcolor{gray}{2.3}}$ & 59.39$_{\textcolor{gray}{0.4}}$ & 40.1(+0.0) \\
          NAdamW-4bit+\textbf{BReD} & 28.98$_{\textcolor{gray}{0.3}}$ & 26.93$_{\textcolor{gray}{1.2}}$ & 49.59$_{\textcolor{gray}{1.1}}$ & 36.00$_{\textcolor{gray}{0.6}}$ & 22.04$_{\textcolor{gray}{0.8}}$ & 60.44$_{\textcolor{gray}{2.4}}$ & 60.01$_{\textcolor{gray}{0.1}}$ & 40.6(+0.5)\\
          NAdamW-3bit+\textbf{BReD} & 28.91$_{\textcolor{gray}{0.2}}$ & 25.93$_{\textcolor{gray}{1.4}}$ & 51.12$_{\textcolor{gray}{1.6}}$ & 35.64$_{\textcolor{gray}{0.7}}$ & 21.96$_{\textcolor{gray}{0.5}}$ & 60.20$_{\textcolor{gray}{2.4}}$ & 59.67$_{\textcolor{gray}{0.6}}$ & 40.5(+0.4)  \\
          \midrule
          SOAP & 29.69$_{\textcolor{gray}{0.1}}$ & 24.93$_{\textcolor{gray}{0.3}}$ & 50.12$_{\textcolor{gray}{1.2}}$ & 36.60$_{\textcolor{gray}{0.7}}$ & 22.47$_{\textcolor{gray}{0.5}}$ & 54.08$_{\textcolor{gray}{8.8}}$ & 59.43$_{\textcolor{gray}{1.1}}$ & 39.6(+0.0) \\
          SOAP-4bit+\textbf{BReD} & 28.97$_{\textcolor{gray}{0.1}}$ & 27.20$_{\textcolor{gray}{0.2}}$ & 50.54$_{\textcolor{gray}{0.9}}$ & 36.02$_{\textcolor{gray}{0.7}}$ & 21.73$_{\textcolor{gray}{0.4}}$ & 58.03$_{\textcolor{gray}{5.1}}$ & 59.36$_{\textcolor{gray}{0.6}}$ & 40.3(+0.7) \\
          SOAP-3bit+\textbf{BReD} & 28.86$_{\textcolor{gray}{0.1}}$ & 26.47$_{\textcolor{gray}{0.8}}$ & 50.70$_{\textcolor{gray}{0.5}}$ & 35.62$_{\textcolor{gray}{0.8}}$ & 22.47$_{\textcolor{gray}{0.8}}$ & 61.04$_{\textcolor{gray}{1.7}}$ & 59.01$_{\textcolor{gray}{1.2}}$ & 40.6(+1.0) \\
          \midrule
          MARS & 29.03$_{\textcolor{gray}{0.1}}$ & 27.20$_{\textcolor{gray}{0.5}}$ & 51.65$_{\textcolor{gray}{1.1}}$ & 36.00$_{\textcolor{gray}{1.0}}$ & 21.56$_{\textcolor{gray}{0.5}}$ & 60.89$_{\textcolor{gray}{2.3}}$ & 59.25$_{\textcolor{gray}{1.0}}$ & 40.8(+0.0) \\
          MARS-4bit+\textbf{BReD} & 28.88$_{\textcolor{gray}{0.1}}$ & 25.20$_{\textcolor{gray}{0.7}}$ & 50.54$_{\textcolor{gray}{0.3}}$ & 35.69$_{\textcolor{gray}{0.3}}$ & 22.58$_{\textcolor{gray}{0.7}}$ & 59.67$_{\textcolor{gray}{1.0}}$ & 59.99$_{\textcolor{gray}{0.3}}$ & 40.4(-0.4)  \\
          MARS-3bit+\textbf{BReD} & 28.59$_{\textcolor{gray}{0.3}}$ & 27.47$_{\textcolor{gray}{1.2}}$ & 51.93$_{\textcolor{gray}{0.5}}$ & 35.91$_{\textcolor{gray}{0.6}}$ & 22.55$_{\textcolor{gray}{1.2}}$ & 61.22$_{\textcolor{gray}{1.0}}$ & 59.49$_{\textcolor{gray}{1.0}}$ & 41.0(+0.2) \\
          \bottomrule
        \end{tabular}
      \end{sc}
    \end{small}
  \end{center}
  \vskip -0.1in
  \vspace{-10pt}
\end{table}

\begin{table*}[h]
  \caption{
  Zero-shot LM evaluation accuracies and validation PPL on 1.1B and 3.4B models.
  Subscripts denote standard deviations over three runs.
  Parentheses report the difference relative to the full-precision optimizer-state baseline of the same optimizer.
  }
  \label{tab:1b_3b_result}
  \setlength{\tabcolsep}{1pt}
  \begin{center}
    \begin{footnotesize}
      \begin{sc}
      \scalebox{0.98}{
        \begin{tabular}{c|c|ccccccc|cc}
          \toprule
          \#Param & Optimizer  & HS & OBQA & WG  & Arc\_e & Arc\_c & BoolQ & PIQA  & Avg. $\uparrow$ & PPL $\downarrow$\\
          \midrule
          \multirow{12}{*}{1.1B} 
          & AdamW    & 43.45$_{\textcolor{gray}{0.1}}$ & 30.80$_{\textcolor{gray}{0.4}}$ & 52.62$_{\textcolor{gray}{0.5}}$ & 45.38$_{\textcolor{gray}{0.4}}$ & 25.66$_{\textcolor{gray}{0.7}}$ & 59.66$_{\textcolor{gray}{0.1}}$ & 68.05$_{\textcolor{gray}{0.7}}$ & 46.5(+0.0) & 10.8(+0.0) \\
          & AdamW-4bit+\textbf{BReD} & 43.05$_{\textcolor{gray}{0.3}}$ & 30.67$_{\textcolor{gray}{1.3}}$ & 52.12$_{\textcolor{gray}{0.2}}$ & 45.50$_{\textcolor{gray}{2.0}}$ & 25.51$_{\textcolor{gray}{0.5}}$ & 59.63$_{\textcolor{gray}{1.8}}$ & 68.03$_{\textcolor{gray}{0.2}}$ & 46.4(-0.1) & 10.9(+0.1) \\
          & AdamW-3bit+\textbf{BReD} & 42.21$_{\textcolor{gray}{2.2}}$ & 30.60$_{\textcolor{gray}{0.5}}$ & 52.78$_{\textcolor{gray}{1.6}}$ & 43.74$_{\textcolor{gray}{2.4}}$ & 24.52$_{\textcolor{gray}{1.1}}$ & 59.14$_{\textcolor{gray}{1.6}}$ & 67.45$_{\textcolor{gray}{1.6}}$ & 45.8(-0.7) & 10.9(+0.1) \\
          \cmidrule{2-11}
          & Muon & 46.01$_{\textcolor{gray}{0.5}}$ & 31.07$_{\textcolor{gray}{0.6}}$ & 52.75$_{\textcolor{gray}{1.1}}$ & 48.23$_{\textcolor{gray}{1.1}}$ & 26.37$_{\textcolor{gray}{0.5}}$ & 59.57$_{\textcolor{gray}{2.7}}$ & 69.10$_{\textcolor{gray}{0.5}}$ & 47.6(+0.0) & 10.3(+0.0) \\
          & Muon-4bit+\textbf{BReD} & 44.46$_{\textcolor{gray}{0.4}}$ & 30.87$_{\textcolor{gray}{1.6}}$ & 53.57$_{\textcolor{gray}{0.3}}$ & 47.28$_{\textcolor{gray}{1.1}}$ & 25.57$_{\textcolor{gray}{0.6}}$ & 58.96$_{\textcolor{gray}{4.2}}$ & 68.93$_{\textcolor{gray}{0.3}}$ & 47.1(-0.5) & 10.6(+0.3) \\
          & Muon-3bit+\textbf{BReD} & 43.08$_{\textcolor{gray}{0.3}}$ & 31.73$_{\textcolor{gray}{0.7}}$ & 51.73$_{\textcolor{gray}{0.7}}$ & 46.09$_{\textcolor{gray}{1.5}}$ & 25.48$_{\textcolor{gray}{0.1}}$ & 60.58$_{\textcolor{gray}{0.5}}$ & 68.19$_{\textcolor{gray}{0.3}}$ & 46.7(-0.9) & 10.9(+0.6) \\
          \midrule
          \multirow{8}{*}{3.4B} 
          & AdamW & 48.01$_{\textcolor{gray}{0.3}}$ & 31.33$_{\textcolor{gray}{0.3}}$ & 53.99$_{\textcolor{gray}{1.8}}$ & 48.45$_{\textcolor{gray}{0.4}}$ & 26.93$_{\textcolor{gray}{0.4}}$ & 56.47$_{\textcolor{gray}{4.7}}$ & 69.98$_{\textcolor{gray}{0.6}}$ & 47.9(+0.0) & 9.7(+0.0) \\
          & AdamW-4bit+\textbf{BReD} & 47.08$_{\textcolor{gray}{0.7}}$ & 31.47$_{\textcolor{gray}{0.7}}$ & 52.80$_{\textcolor{gray}{1.8}}$ & 47.53$_{\textcolor{gray}{1.2}}$ & 26.08$_{\textcolor{gray}{0.2}}$ & 59.25$_{\textcolor{gray}{1.0}}$ & 69.24$_{\textcolor{gray}{0.2}}$ & 47.6(-0.3) & 9.8(+0.1) \\
          \cmidrule{2-11}
          & Muon & 48.37$_{\textcolor{gray}{0.2}}$ & 31.53$_{\textcolor{gray}{0.3}}$ & 54.91$_{\textcolor{gray}{1.3}}$ & 49.17$_{\textcolor{gray}{0.4}}$ & 26.48$_{\textcolor{gray}{0.1}}$ & 57.50$_{\textcolor{gray}{2.8}}$ & 69.68$_{\textcolor{gray}{0.5}}$ & 48.2(+0.0) & 9.6(+0.0) \\
          & Muon-4bit+\textbf{BReD} & 46.31$_{\textcolor{gray}{0.0}}$ & 30.47$_{\textcolor{gray}{0.3}}$ & 54.96$_{\textcolor{gray}{0.9}}$ & 47.11$_{\textcolor{gray}{1.2}}$ & 26.23$_{\textcolor{gray}{0.4}}$ & 59.77$_{\textcolor{gray}{1.0}}$ & 68.74$_{\textcolor{gray}{0.4}}$ & 47.7(-0.5) & 9.8(+0.2) \\
          \bottomrule
        \end{tabular}
        }
      \end{sc}
    \end{footnotesize}
  \end{center}
  \vskip -0.1in
\end{table*}

\newpage
\subsection{3-bit results}
\label{sec:apd_3bit_results}
\begin{table}[h]
\centering
\small
\setlength{\tabcolsep}{3.8pt}
\caption{
Aggressive 3-bit BReD compression. 
For pretraining, Acc. denotes average zero-shot accuracy; for supervised fine-tuning, Avg. denotes average downstream score. 
Parentheses denote changes relative to the corresponding full-precision optimizer-state baseline.
}
\label{tab:three-bit}
\begin{tabular}{llllcccc}
\toprule
\multirow{2}{*}{Stage}
& \multirow{2}{*}{Architecture}
& \multirow{2}{*}{Size}
& \multirow{2}{*}{Optimizer}
& \multicolumn{2}{c}{Acc. $\uparrow$}
& \multicolumn{2}{c}{PPL $\downarrow$} \\
\cmidrule(lr){5-6} \cmidrule(lr){7-8}
& & & & FP32-state & 3-bit BReD & FP32-state & 3-bit BReD \\
\midrule
Pretrain & LLaMA-style & 120M & AdamW  & 40.4 & 40.6 (+0.2) & 18.3 & 19.0 (+0.7) \\
Pretrain & LLaMA-style & 120M & Muon   & 40.9 & 40.9 (+0.0) & 17.8 & 18.5 (+0.7) \\
Pretrain & LLaMA-style & 120M & NAdamW & 40.1 & 40.5 (+0.4) & 18.9 & 19.0 (+0.1) \\
Pretrain & LLaMA-style & 120M & SOAP   & 39.6 & 40.6 (+1.0) & 17.6 & 18.8 (+1.2) \\
Pretrain & LLaMA-style & 120M & MARS   & 40.8 & 41.0 (+0.2) & 18.7 & 19.4 (+0.7) \\
\midrule
Pretrain & LLaMA-style & 1.1B & AdamW  & 46.5 & 45.8 (-0.7) & 10.8 & 10.9 (+0.1) \\
Pretrain & LLaMA-style & 1.1B & Muon   & 47.6 & 46.7 (-0.9) & 10.3 & 10.9 (+0.6) \\
\midrule
SFT & Qwen2.5 & 7B & AdamW & 54.2 & 55.0 (+0.8) & 4.6 & 4.7 (+0.1) \\
SFT & Qwen2.5 & 7B & Muon  & 55.2 & 54.4 (-0.8) & 4.6 & 4.5 (-0.1) \\
\bottomrule
\end{tabular}
\end{table}

\subsection{Higher EMA decay}
\label{app:high_beta}

We additionally evaluate 120M AdamW pretraining with
$\beta_2=0.999$, changing only $\beta_2$ from the default
training recipe.
FP32-state AdamW and 4-bit BReD use the same $\beta_2$
within each comparison.
Table~\ref{tab:high_beta} shows that BReD remains stable
at $\beta_2=0.999$, with a PPL increase of $0.7$.
This is a single paired run, rather than a systematic
high-$\beta$ sweep; we do not infer a cross-$\beta_2$
trend from these results.

\begin{table}[h]
\centering
\small
\caption{
Validation PPL for 120M AdamW at two EMA decay factors.
The $\beta_2=0.95$ results are means over three runs;
the $\beta_2=0.999$ results are from one paired run.
$\Delta$PPL is BReD minus FP32-state at the same $\beta_2$.
}
\label{tab:high_beta}
\begin{tabular}{lrrr}
\toprule
$\beta_2$ & FP32-state & 4-bit BReD & $\Delta$PPL \\
\midrule
0.95  & 18.3 & 18.7 & +0.4 \\
0.999 & 19.4 & 20.1 & +0.7 \\
\bottomrule
\end{tabular}
\end{table}

\subsection{Additional quantization format}
\label{app:mxint4}

We evaluate MXINT4 in the same 120M AdamW setup,
changing only the quantization format used for optimizer-state
storage.
As shown in Table~\ref{tab:mxint4}, 4-bit BReD with MXINT4
remains convergent, reaching a validation PPL of 18.9,
compared with 18.7 for MXFP4.

\begin{table}[h]
\centering
\small
\caption{
Validation PPL for 120M AdamW with different state-storage
formats. All values are means over three runs.
}
\label{tab:mxint4}
\begin{tabular}{llr}
\toprule
State storage & Format & Validation PPL \\
\midrule
FP32-state & FP32 & 18.3 \\
4-bit BReD & MXFP4 & 18.7 \\
4-bit BReD & MXINT4 & 18.9 \\
\bottomrule
\end{tabular}
\end{table}

\subsection{SOLO checks}
\label{app:solo}

We evaluate the author-released SOLO implementation in our
120M from-scratch pretraining setup.
For each precision setting, we retain SOLO's official
quantization configuration and vary the AdamW betas as listed
in Table~\ref{tab:solo}.
The model, data, learning-rate schedule, weight decay,
batch size, and seed are fixed across runs.
All SOLO runs use quantile $0.1$ and block size $128$.

\begin{table}[h]
\centering
\small
\caption{
SOLO checks in our 120M pretraining setup.
Precision denotes first-/second-moment bitwidths.
Each entry reports one run.
}
\label{tab:solo}
\begin{tabular}{llcl}
\toprule
Beta setting & Precision & $(\beta_1,\beta_2)$ & Result \\
\midrule
Original AdamW     & 4/2-bit & $(0.9,0.95)$  & Diverged \\
SOLO from-scratch  & 4/2-bit & $(0.3,0.95)$  & Diverged \\
SOLO from-scratch  & 2/2-bit & $(0.1,0.95)$  & Diverged \\
Author example    & 4/2-bit & $(0.8,0.999)$ & Diverged \\
\bottomrule
\end{tabular}
\end{table}

For reference, 3/3-bit BReD reaches validation PPL 19.0
in the same pretraining setup with the original AdamW
hyperparameters.
These limited checks do not imply that SOLO fails in its
reported settings, nor do they identify which differences
in quantization or optimizer dynamics caused the observed
failures.

\subsection{Small-scale Shampoo check}
\label{app:shampoo}

We additionally evaluate 120M Shampoo with 4-bit
preconditioner states, keeping the model, data, optimizer
update, and hyperparameters unchanged across configurations.
As shown in Table~\ref{tab:shampoo}, BReD completes training
with validation PPL 18.5, compared with 18.3 for FP32-state
Shampoo, whereas NR and SR diverge.

\begin{table}[h]
\centering
\small
\caption{
Validation PPL for the 120M Shampoo check.
Low-bit variants quantize the preconditioner states to 4 bits.
Each configuration reports one run.
}
\label{tab:shampoo}
\begin{tabular}{lr}
\toprule
Method & Validation PPL \\
\midrule
FP32-state Shampoo & 18.3 \\
Shampoo + NR       & Diverged \\
Shampoo + SR       & Diverged \\
Shampoo + BReD     & 18.5 \\
\bottomrule
\end{tabular}
\end{table}

This single-run check does not support a statistical comparison
or replace large-scale validation.
It also does not establish superiority over dedicated methods
such as 4-bit Shampoo~\citep{wang20244} or 4-bit Preconditioned
Optimization~\citep{li2025memory}, which are not evaluated here.

\subsection{memory reduction}
\label{sec:memory}
\begin{table}[h]
\centering
\small
\caption{
Optimizer-state memory reduction with 4-bit BReD. 
% Memory is measured for the packed optimizer states used during training. 
Reduction is relative to the corresponding full-precision optimizer-state baseline.
}
\label{tab:memory-reduction}
\begin{tabular}{llccc}
\toprule
Scale & Optimizer & FP32-state & 4-bit BReD & Reduction \\
\midrule
120M & AdamW  & 0.90GB & 0.19GB & 78.9\% \\
120M & Muon   & 0.64GB & 0.41GB & 35.9\% \\
120M & NAdamW & 0.90GB & 0.19GB & 78.9\% \\
120M & SOAP   & 1.66GB & 0.55GB & 66.9\% \\
120M & MARS   & 1.36GB & 0.64GB & 52.9\% \\
\midrule
1.1B & AdamW & 8.20GB & 1.18GB & 85.6\% \\
1.1B & Muon  & 4.59GB & 1.47GB & 68.0\% \\
\midrule
3.4B & AdamW & 25.53GB & 3.58GB & 86.0\% \\
3.4B & Muon  & 13.53GB & 3.17GB & 76.6\% \\
\bottomrule
\end{tabular}
\end{table}

\section{Implementation note.}
\paragraph{PRNG}
We use stateless PRNG (e.g., Philox/Threefry-style) so that each $u$ can be generated independently in parallel from its key.
This avoids maintaining PRNG states across threads and guarantees bitwise reproducibility under re-ordering.

\paragraph{Block-wise SD.}
For SD, we similarly reuse $u^{(k)}$ to quantize all $\alpha(x_i)$ within the block and replay the same $u^{(k)}$
during dequantization.
This keeps the coordinate-wise error unbiased while greatly reducing PRNG calls.

\section{Proof}
\label{sec:proof}
\begin{lemma}[mean induces persistent drift under EMA recursion]
Assume the injected errors satisfy $\mathbb{E}[\epsilon_t]=b$.
For the recursion $e_{t+1}=\beta e_t+\epsilon_{t+1}$ with $0<\beta<1$, we have
\begin{equation}
\mathbb{E}[e_t] = \beta^t \mathbb{E}[e_0] + \frac{1-\beta^t}{1-\beta}\,b,
\end{equation}
and therefore the mean error converges to
\begin{equation}
\lim_{t\to\infty}\mathbb{E}[e_t] = \frac{b}{1-\beta}.
\end{equation}
In particular, even a small nonzero bias $b$ yields a persistent drift whose magnitude scales as $\mathcal{O}\!\left(\frac{\|b\|}{1-\beta}\right)$ when $\beta\to 1$.
\end{lemma}

\paragraph{Proof.}
Unrolling $e_t=\beta^t e_0 + \sum_{k=1}^t \beta^{t-k}\epsilon_k$ and taking expectation gives the result.

\begin{theorem}[Steady-state second moment under quantized EMA]
Assume $\epsilon_t=b+\xi_t$, where $\{\xi_t\}_{t\geq 1}$ are independent with $\mathbb{E}[\xi_t]=0$ and $\mathbb{E}\|\xi_t\|_F^2=\sigma^2$.
Assume also that $e_0$ has finite second moment and is independent of $\{\xi_t\}_{t\geq 1}$.
For $0<\beta<1$, the state error satisfies
\begin{equation}
\lim_{t\to\infty}\mathbb{E}\|e_t\|_F^2
=
\underbrace{\frac{\|b\|_F^2}{(1-\beta)^2}}_{\text{mean (bias) term}}
+
\underbrace{\frac{\sigma^2}{1-\beta^2}}_{\text{variance (noise) term}}.
\end{equation}
\end{theorem}

\paragraph{Proof.}
Unrolling Eq.~\eqref{eq:ema_error_recursion} gives $e_t=\beta^t e_0+\sum_{k=1}^t\beta^{t-k}\epsilon_k$.
Taking expectations yields $\lim_{t\to\infty}\mathbb{E}[e_t]=b/(1-\beta)$.
Writing $\epsilon_k=b+\xi_k$ with $\mathbb{E}[\xi_k]=0$ and using independence cancels cross terms, leading to Eq.~\eqref{eq:ema_second_moment}.

\paragraph{Proof (SD unbiasedness and variance).}
\label{proof:sd}
Let $v := \{\alpha + u\}$ denote the fractional part of $\alpha+u$.
Since $u\sim\mathrm{Unif}(0,1)$, we also have $v\sim\mathrm{Unif}(0,1)$.
Moreover,
\[
\hat{\alpha} - \alpha
= c_{\text{sd}}(\alpha;u) - u + \tfrac12 - \alpha
= \tfrac12 - \{\alpha + u\}
= \tfrac12 - v.
\]
Hence $\hat{\alpha}-\alpha \sim \mathrm{Unif}(-1/2,\,1/2)$, which implies
\[
\mathbb{E}[\hat{\alpha}-\alpha\mid x]=0,\qquad
\mathrm{Var}(\hat{\alpha}-\alpha\mid x)=1/12.
\]
Multiplying by $\Delta(x)$ yields
\[
\mathbb{E}[\epsilon_{\text{elem}}\mid x]=0,\qquad
\mathrm{Var}(\epsilon_{\text{elem}}\mid x)=\Delta(x)^2/12.
\]

\subsection{Rounding-error tails of SR and SD}
\label{app:rounding_tails}

Under the local non-clipped interval model, let
$r=(\hat{x}-x)/\Delta(x)$.
For a fixed $\alpha$, SR gives errors $1-\alpha$ and $-\alpha$
with probabilities $\alpha$ and $1-\alpha$, respectively,
whereas SD gives $r_{\mathrm{SD}}\sim\mathrm{Unif}(-1/2,1/2)$.
Under the illustrative assumption $\alpha\sim\mathrm{Unif}(0,1)$,
\[
\Pr(|r_{\mathrm{SR}}|>\tau)=(1-\tau)^2
\quad (0\leq\tau\leq1),
\qquad
\Pr(|r_{\mathrm{SD}}|>\tau)=0
\quad (\tau>1/2).
\]
Thus, unbiased SR can produce larger rounding errors than SD.
This offers a possible explanation for the observed SR failures,
but does not establish a training-divergence condition under
Theorem~\ref{thm:ema_second_moment}.

\newpage
\section{Additional In-Training Diagnostics}
\label{sec:fig_plot}
\begin{figure}[h]
    \centering
    \begin{subfigure}[t]{0.49\textwidth}
        \centering
        \makebox[\linewidth][c]{
        \includegraphics[height=\subfigH,
        width = \textwidth,
        keepaspectratio]{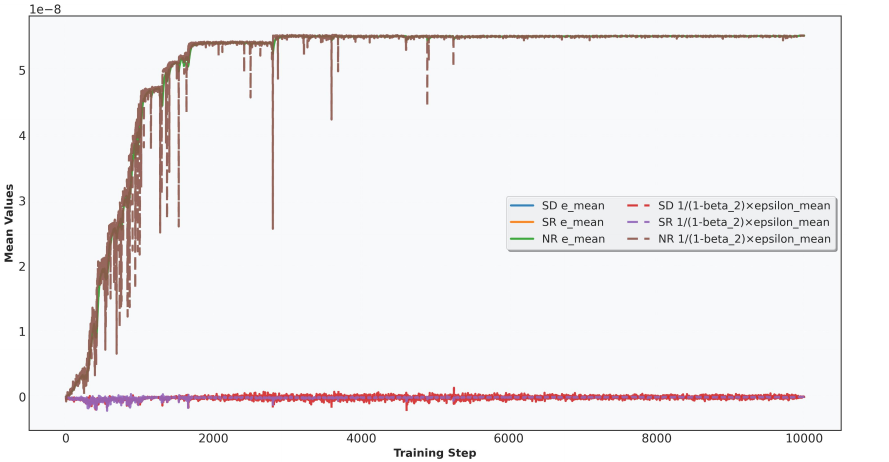}}
        \caption{Layer 1.attn.attn, show the $\mathbb{E}[e_t]$ and $\mathbb{E}[\epsilon_t]/(1-\beta_2)$.}
    \end{subfigure}
    \hfill
    \begin{subfigure}[t]{0.49\textwidth}
        \centering
        \makebox[\linewidth][c]{
        \includegraphics[height=\subfigH,
        width = \textwidth,
        keepaspectratio]{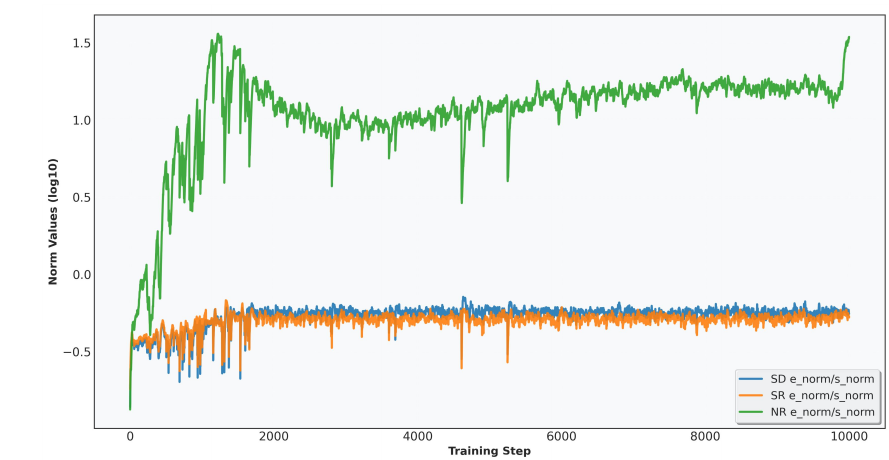}}
        \caption{Layer 1.attn.attn, error accumulates quickly.}
    \end{subfigure}

    \vspace{0.2em}

    \begin{subfigure}[t]{0.49\textwidth}
        \centering
        \makebox[\linewidth][c]{
        \includegraphics[height=\subfigH,
        width = \textwidth,
        keepaspectratio]{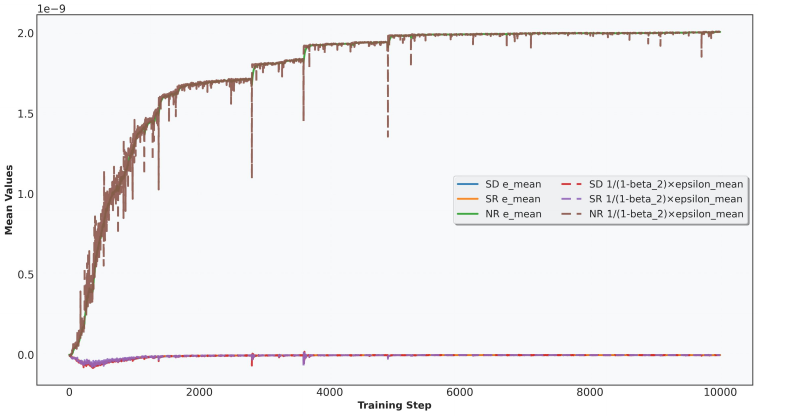}}
        \caption{Layer 5.mlp.swiglu.w1, show the $\mathbb{E}[e_t]$ and $\mathbb{E}[\epsilon_t]/(1-\beta_2)$.}
    \end{subfigure}
    \hfill
    \begin{subfigure}[t]{0.49\textwidth}
        \centering
        \makebox[\linewidth][c]{
        \includegraphics[height=\subfigH,
        width = \textwidth,
        keepaspectratio]{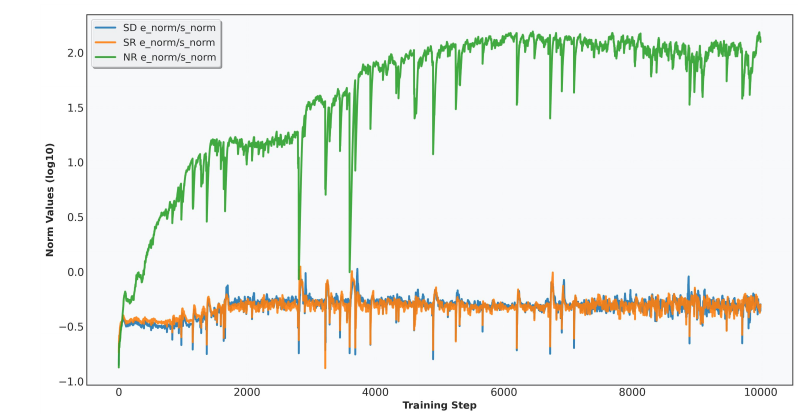}}
        \caption{Layer 5.mlp.swiglu.w1, error accumulate quickly.}
    \end{subfigure}

    \caption{\textbf{Additional quantized-EMA diagnostics for two layers:} Panels (a) and (c) compare the measured mean accumulated error $\mathbb{E}[e_t]$ with the predicted amplified injected error $\mathbb{E}[\epsilon_t]/(1-\beta_2)$. Panels (b) and (d) show the relative error norm over training. Across layers, nearest rounding accumulates large error, whereas stochastic rounding and subtractive dithering maintain bounded relative errors in this diagnostic.
}
\end{figure}

%%%%%%%%%%%%%%%%%%%%%%%%%%%%%%%%%%%%%%%%%%%%%%%%%%%%%%%%%%%%

% \newpage
% \input{checklist.tex}

\end{document}